\pdfoutput=1
\documentclass[11pt]{article}
\usepackage{arabtex}
\usepackage{utf8}
\setcode{utf8}
\usepackage{acl}
\usepackage{times}
\usepackage{latexsym}
\usepackage[T1]{fontenc}
\usepackage[utf8]{inputenc}
\usepackage[colorinlistoftodos,prependcaption,textsize=tiny]{todonotes}
\usepackage{microtype}
\usepackage{inconsolata}
\usepackage{microtype}
\usepackage{inconsolata}
\usepackage{algorithm}
\usepackage{algpseudocode}
\usepackage{xspace}
\newcommand{\Ar}[1]{{\scriptsize\<#1>}}
\usepackage{multirow}

\newcommand{\TrAr}[1]{\arabtrue\transfalse {\small \Ar{#1}} / \arabfalse\transtrue \RL{#1}\arabtrue\transfalse }
\title{\quran: Morphologically Annotated Quranic Corpus}

\author{Diyam Akra \\
  Birzeit University \\
  \texttt{dakra@birzeit.edu} \\\And
 Tymaa Hammouda\\
Birzeit University \\
\texttt{thammouda@birzeit.edu} 
\\\And
 Mustafa Jarrar\\
Birzeit University \\
\texttt{mjarrar@birzeit.edu} \\}

\newcommand{\quran}{\cal{\textit{QuranMorph}}\xspace}

\begin{document}
\maketitle
\begin{abstract}

We present the \quran corpus, a morphologically annotated corpus for the Quran (77,429 tokens). Each token in the \quran was manually lemmatized and tagged with its part-of-speech by three expert linguists. The lemmatization process utilized lemmas from Qabas, an Arabic lexicographic database linked with 110 lexicons and corpora of 2 million tokens. The part-of-speech tagging was performed using the fine-grained SAMA/Qabas tagset, which encompasses 40 tags. As shown in this paper, this rich lemmatization and POS tagset enabled the \quran corpus to be inter-linked with many linguistic resources. The corpus is open-source and publicly available as part of the SinaLab resources at ({\small \url{https://sina.birzeit.edu/quran}}).
\end{abstract}

\section{Introduction}
The Arabic language exists in three forms~\cite{J21}: Classical Arabic is the Arabic used between the 6th and the 15th centuries, including the  Quran and the prophetic traditions; Modern Standard Arabic (MSA), is the language used in official and formal contexts nowadays; and the Dialectal Arabic includes the diverse spoken varieties across Arab countries. The development of computational resources for Classical Arabic aligns with the growing interest in advancing resources for both MSA and dialects~\cite{DH21}.

This paper introduces a lemmatized and POS-tagged Quranic corpus called \quran. The lemmatization and POS-tagging were carried out manually by three expert linguistic annotators. The final annotated corpus contains 77,429 words.

\section{Related Work}
The Quranic Arabic Corpus is the only available morphologically annotated corpora for Classical Arabic \citep{dukes2010morphological}. The POS tagset used to annotate this corpus consists of 44 tags, which were designed to cover traditional Arabic grammar, and the Quran in particular. The challenge is that these POS annotations cannot be easily integrated with the available corpora. Additionally, the lemmatization of the Quranic Corpus did not also utilize lemmas from known lexicographic resources, such as the SAMA database \cite{Graff2009}. 
Consequently, using the Quranic Corpus alongside other resources as a training dataset is challenging due to its entirely different lemmas and POS tagset compared to other resources.

There are other related morphologically annotated resources available for Modern Standard Arabic (MSA). These include the Prague Arabic Dependency Treebank (Ar-PADT) \cite{Smrz2008} and LDC’s Penn Arabic Treebank (PATB) \cite{Maamouri2009}.

At the dialectal level, several morphologically annotated resources are available for various Arabic dialects. These include the Egyptian Arabic Treebank (ARZATB) \cite{Maamouri2014}, a morphologically annotated treebank for Egyptian Arabic; Curras \cite{JHRAZ17}, a morphologically annotated corpus for Palestinian Arabic; Baladi \cite{EJHZ22}, a corpus for Lebanese Arabic; Nabra \cite{ANMFTM23}, a corpus for Syrian Arabic; Gumar \cite{Khalifa2018}, a corpus for Emirati Arabic; and Lisan \cite{JZHNW23}, which provides morphologically annotated corpora for Libyan, Yemeni, Iraqi, and Sudanese Arabic.

All of these corpora are annotated using the lemmas and tagsets in either SAMA or Qabas. The LDC’s SAMA \cite{Graff2009} is a stem database containing stems, their lemmas, and compatible affixes. Qabas \cite{JH24} is an Arabic lexicographic database that consists of about $60K$ lemmas, linked with $110$ Arabic lexicons and 2 million tokens corpora \cite{J21,JA19,JZAA18}. Qabas is also linked with other resources for synonymy \cite{GJJB23,JKKS21}, Word Sense Disambiguation \cite{JMHK23,HJ21b}, among other resources. 
The idea is that corpora annotated with similar or closely related lemmas and tagsets are easier to unify and utilize for joint training.
 
\section{The Quran Corpus}
As mentioned earlier, the only valuable corpus for the Quran was developed in $2011$ by \citet{dukes2010morphological}. This corpus  of $128,219$ forms (clitics) extracted from the Quran, each is annotated with a set of morphological features. To ensure compatibility with this corpus, we decided to adopt the same segmentation and tokenization methodology. The corpus includes a location feature, which specifies the position of each clitic in terms of chapter, verse, word, and its position within the word. For instance, a location of (1:1:1:1) refers to the first clitic in the first word of the first verse in the first chapter. In alignment with this, we adopted the same tokenization but focused solely on words rather than clitics. In other words, we combined clitics to form complete words. As a result, each chapter consists of verses, and each verse consists of words, represented by the location format (1:1:1). As shown in Table \ref{table:quran_stat} the total number of words in the Quran is $77,429$ (which are $19,009$ unique words) in $6,235$ verses. We then annotated these words with their corresponding part-of-speech tags and lemmas, as explained in the following sections.

Notably, our annotations can be seamlessly integrated with those in the Quranic Corpus using the location format (1:1:1), allowing for the combined use of both corpora.

\section{Annotation Methodology}
\label{sec:methodology}
This section presents the lemmatization and POS tagging of the Quran Corpus. 

\begin{figure*}[!ht]
  \centering
  \includegraphics[width=1\textwidth]{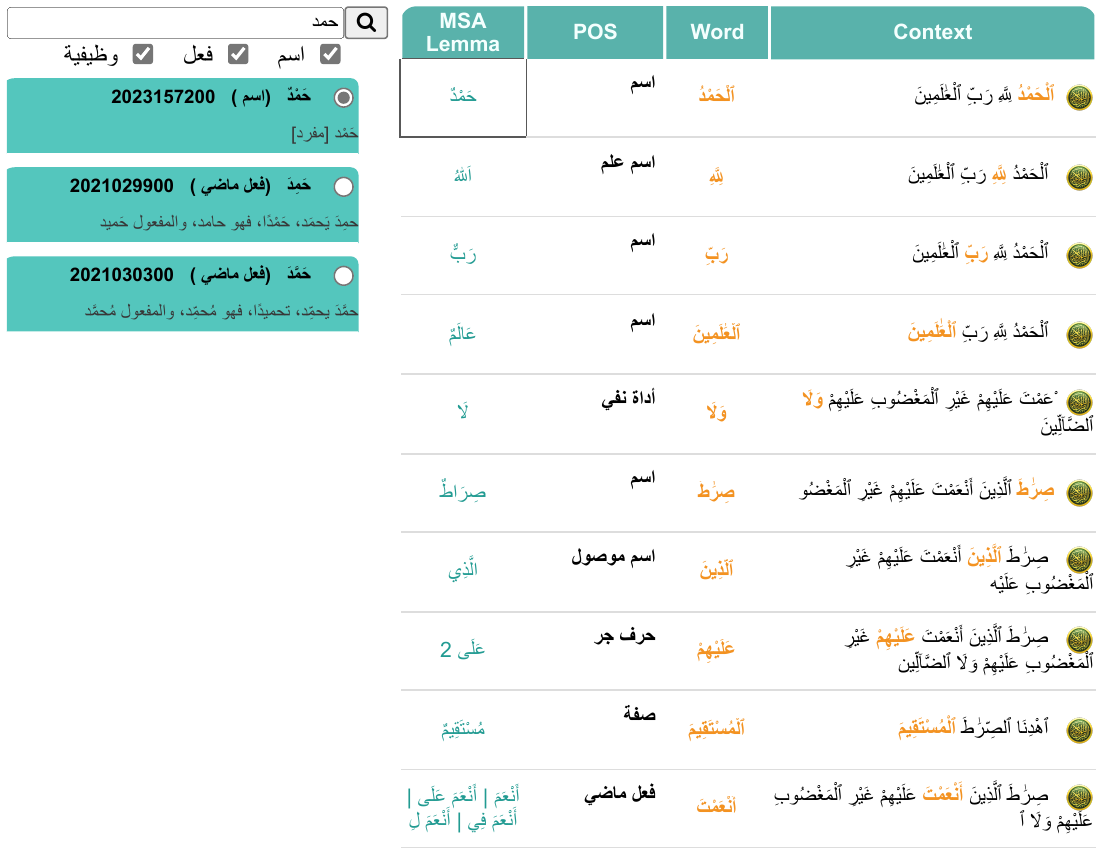}
  \caption{Screenshot of the Tawseem Annotation Tool with example annotations}
  \label{fig:tawseem_portal}
\end{figure*}

The corpus was uploaded to the Tawseem Web Annotation tool (See the screenshot in Figure \ref{fig:tawseem_portal}). The corpus was then assigned to three expert linguists for annotation, which they completed in approximately $400$ working hours at a rate of $5$ USD per hour. Tawseem’s smart features streamline the annotation process by automating and validating tasks, significantly boosting both quality and productivity.

As shown in the figure, the first column represents the context (i.e., verse) where the target word to be annotated is highlighted. The third and fourth columns are the POS and lemma assigned by the linguists with the assistance of the tool.
Tawseem integrates with the Alma morphological tagger \cite{JAH24} from SinaTools \cite{HJK24}, allowing it to suggest annotations (lemmas and POS tags) for review and confirmation by annotators.

\subsection{Lemmatization}
To confirm the lemma suggested by the Tawseem tool the annotator has to carefully review the context (i.e., verse) containing the target word to ensure an accurate understanding of its meaning. This step is crucial for selecting the most appropriate lemma, taking into account the word's syntactic role and its semantic relationship with the surrounding text. 
If the suggested lemma is not appropriate, the annotator searches Qabas for other lemmas using the search bar located at the top left of Figure \ref{fig:tawseem_portal}. If no suitable lemma is found, a new Lemma entry is added to Qabas, complete with relevant linguistic and morphological details.

The idea of using Qabas lemmas to annotate the Quran corpus stems from their unique linkage to lemmas in 110 lexicons. Additionally, Qabas lemmas are used to annotate various MSA and dialectal corpora, totaling around 2 million tokens. This integration allows the Quran corpus to be utilized alongside other corpora for training purposes, offering a more comprehensive resource for Classical, MSA, and dialectal Arabic. 

The online version of Qabas\footnote{Qabas: \url{https://sina.birzeit.edu/qabas}} allows users to search for a lemma and view the contexts in which this lemma appears in the Quran and other corpora.

\subsection{POS Tagging}
Similarly, to confirm the POS tag suggested by the Tawseem tool, the annotator must carefully review the entire verse containing the target word to ensure an accurate understanding of its context. This step is essential for assigning the most appropriate POS tag. If the suggested POS tag is unsuitable, the annotator can select an alternative from the drop-down Qabas tagset, which includes the same 40 POS tags used in LDC’s SAMA \cite{Graff2009}.

To sum up, the Table \ref{table:quran_stat} presents statistics about the corpora.

\begin{table}
 \centering
  \begin{tabular}{|l|r|}      \hline
   Words & 77,429\\      \hline
    Unique Words &  19009 \\      \hline
      Verses & 6,235  \\      \hline
      Unique Nouns &   9,789\\      \hline
      Unique Verbs &   8,265\\      \hline
     Unique Functional Words &   1,036\\      \hline
     Unique Lemmas &  4,616 \\      \hline
       Unique Nouns Lemmas &   3,057\\      \hline
      Unique Verbs Lemmas &   1,479\\      \hline
     Unique Functional Words Lemmas &   173\\      \hline
\end{tabular} 
 \caption{Statistics about our \quran Corpus}
  \label{table:quran_stat}
\end{table}

\section{Discussion}
Due to the unique nature of the Quran, annotators encounter distinct challenges that can be grouped into the following categories:

 \begin{itemize}
 
        \item Difficulties arose due to the Quranic script (\TrAr{الرسم القرآني}), which is not same as standard Arabic orthography, making the annotation process more complex. These differences include variations in spelling, diacritization, and letter forms, requiring annotators to carefully analyze each word within its contextual and historical linguistic framework. For example, the verse (\Ar{الذين يؤمنون بالغيب و يقيمون الصلاة و مما رزقناهم ينفقون}) appears as (\Ar{ٱلَّذِينَ يُؤۡمِنُونَ بِٱلۡغَيۡبِ وَيُقِيمُونَ ٱلصَّلَوٰةَ وَمِمَّا رَزَقۡنَٰهُمۡ يُنفِقُونَ}) in Quranic script.
        
        \item Annotators often had to consult Quranic exegesis (\TrAr{تفسير القرآن}) to accurately determine the most suitable lemma for a target word. This step is necessary because Quranic words, while often resembling standard Arabic vocabulary, frequently carry context-dependent meanings that differ from their conventional usage. For instance, in the verse: (\Ar{ثُمَّ بَدَّلْنَا مَكَانَ السَّيِّئَةِ الْحَسَنَةَ حَتَّىٰ عَفَوا})-Quran 7:95, the term (\TrAr{عَفَوا}) does not mean forgiveness or pardon as it does in Modern Standard Arabic. Instead, within this Quranic context, it signifies an increase or growth.
        
        \item The same word can appear in diverse contexts, leading to variations in its meaning and associated lemma. Consequently, annotators must carefully analyze grammatical conjugations and semantic nuances to accurately determine the correct lemma. For example, in the Quran, the word (\TrAr{ضرب}) carries different meanings depending on the context: In the verse (\Ar{أَلَمْ تَرَ كَيْفَ ضَرَبَ اللَّهُ مَثَلًا كَلِمَةً طَيِّبَةً كَشَجَرَةٍ طَيِّبَةٍ})-Quran 14:24, the verb (\TrAr{ضَرَبَ}) means (to provide an example) or (to illustrate a concept). In contrast, in the verse (\Ar{وَإِذَا ضَرَبْتُمْ فِي الْأَرْضِ فَلَيْسَ عَلَيْكُمْ جُنَاحٌ})-Quran 4:101, (\TrAr{ضَرَبْتُمْ}) conveys (to travel) or (to journey through the land).
    \end{itemize}

\section{Conclusion}
This paper presents the \quran a morphologically annotated corpus for the Quran (77,429 tokens). Each token in the \quran is lemmatized and tagged with its part of speech. The lemmatization process utilized the lemmas from Qabas. The corpus was also tagged with parts of speech using the SAMA/Qabas tagset. 

\section{Acknowledgment}
We extend our thanks to all the annotators who contributed to the annotation process, especially Shimaa Hamayel, Nagham Idrees, and Shahed Khader.

\bibliography{MyReferences,custom}

\end{document}